\documentclass[sigconf, authorversion]{acmart} 
\usepackage{amsmath}
\DeclareMathOperator*{\argmax}{arg\,max}
\DeclareMathOperator*{\argmin}{arg\,min}
\usepackage{multirow}
\usepackage{makecell}
\usepackage{float}
\newcommand{\f}[1]{\textbf{#1}}
\AtBeginDocument{%
  }

\setcopyright{acmlicensed}
\copyrightyear{2025}
\acmYear{2025}
\acmConference[CIKM’25]{The 34th ACM International Conference on Information and Knowledge Management}{November 10–-14, 2025}{Seoul}

\acmSubmissionID{1}

\begin{document}

\title{Decision by Supervised Learning with Deep Ensembles: \\ A Practical Framework for Robust Portfolio Optimization}

\author{Juhyeong Kim}
\email{juhyeong.kim@miraeasset.com}
\email{nonconvexopt@gmail.com}
\affiliation{
  \institution{Mirae Asset Global Investments}
  \institution{AI Quant Lab, MODULABS}
  \city{Seoul}
  \country{Republic of Korea}
}

\author{Sungyoon Choi}
\email{sungyoon.choi90@gmail.com}
\affiliation{
  \institution{Mirae Asset Global Investments}
  \city{Seoul}
  \country{Republic of Korea}
}

\author{Youngbin Lee}
\email{youngandbin@elicer.com}
\affiliation{
  \institution{Elice}
  \institution{AI Quant Lab, MODULABS}
  \city{Seoul}
  \country{Republic of Korea}
}

\author{Yejin Kim}
\email{yejin.kim.ds@meritz.com}
\affiliation{
  \institution{Meritz Fire \& Marine Insurance}
  \institution{AI Quant Lab, MODULABS}
  \city{Seoul}
  \country{Republic of Korea}
}

\author{Yongmin Choi}
\email{fudiso7@gmail.com}
\affiliation{
  \city{Seoul}
  \country{Republic of Korea}
}

\author{Yongjae Lee}
\email{yongjaelee@unist.ac.kr}
\affiliation{
  \institution{Ulsan National Institute of Science and Technology}
  \city{Ulsan}
  \country{Republic of Korea}
}

\renewcommand{\shortauthors}{Kim et al.}


\begin{abstract}
We propose Decision by Supervised Learning (DSL), a practical framework for robust portfolio optimization. DSL reframes portfolio construction as a supervised learning problem: models are trained to predict optimal portfolio weights, using cross-entropy loss and portfolios constructed by maximizing the Sharpe or Sortino ratio. To further enhance stability and reliability, DSL employs Deep Ensemble methods, substantially reducing variance in portfolio allocations. Through comprehensive backtesting across diverse market universes and neural architectures, shows superior performance compared to both traditional strategies and leading machine learning-based methods, including Prediction-Focused Learning and End-to-End Learning. We show that increasing the ensemble size leads to higher median returns and more stable risk-adjusted performance. The code is available at \url{https://github.com/DSLwDE/DSLwDE}.
\end{abstract}



\received{6 August 2025}

\begin{teaserfigure}
  \centering
  \includegraphics[width=\textwidth]{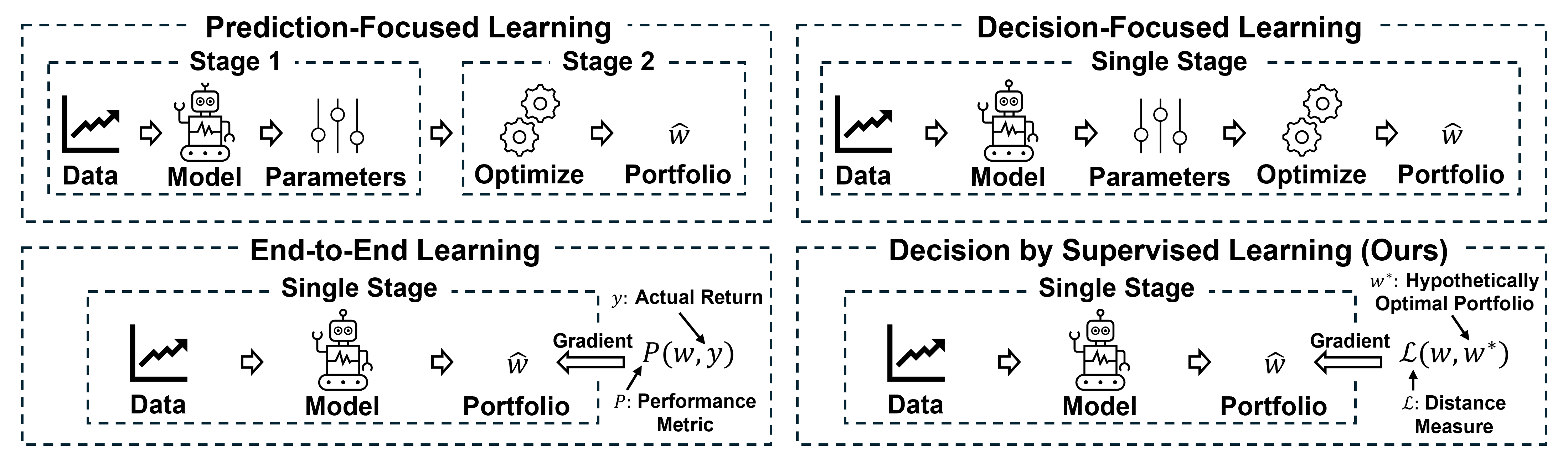}
  \caption{Graphical illustrations of Machine Learning-based Portfolio Optimization methods}
  \Description{This is a graphical illustrations of Prediction-Focused Learning, Decision-Focused Learning, End-to-End Learning and Decision by Supervised Learning}
\end{teaserfigure}

\maketitle

\section{Introduction}
Machine learning has become an essential tool for portfolio optimization, with deep learning methods enabling more adaptive and data-driven investment strategies \cite{lee2023overview}. Among recent developments, supervised learning and end-to-end optimization have been especially influential, offering direct ways to link financial objectives with model training. For instance, \citet{Zhang_2020} demonstrated how the Sharpe ratio—a widely used performance metric \cite{RePEc:ucp:jnlbus:v:39:y:1965:p:119}—can be embedded directly into the loss function of neural networks. This approach, representative of the "Decision-Focused" or "End-to-End" Learning paradigm, seeks to optimize the ultimate investment outcome rather than mere predictive accuracy.

However, optimizing non-convex objectives like the Sharpe ratio introduces significant practical challenges. The optimization landscape is highly non-linear and prone to local minima, making training unstable and results unreliable. Moreover, deep learning models, by their nature, are sensitive to initialization and hyperparameter choices, which can amplify these instabilities. Given these limitations in end-to-end Learning, there is a growing need for robust, stable, and scalable frameworks that can bridge the gap between theoretical advances and practical portfolio construction.

A promising solution is the use of Deep Ensembles, which aggregate the outputs of multiple independently trained neural networks to reduce predictive variance and enhance reliability \cite{gupta2022ensembles}. Ensemble techniques have been shown to improve both accuracy and calibration in various domains, and are particularly valuable in finance, where stability and risk management are paramount.

In this paper, we propose Decision by Supervised Learning (DSL): a practical framework for robust portfolio optimization. DSL reframes portfolio construction as a supervised learning problem, training models to predict optimal portfolio weights directly using convex surrogate losses—such as cross-entropy—rather than non-convex financial metrics. By integrating Deep Ensemble methods, DSL achieves significantly improved stability and out-of-sample performance, providing a robust alternative to both classical and recent machine learning-based approaches.

Our main contributions are as follows:

\begin{itemize}

\item{We introduce the DSL framework, which combines supervised learning with Deep Ensembles for portfolio optimization, supporting a wide range of models and objectives.}

\item{We provide a thorough empirical evaluation, demonstrating that shows superior performance compared to traditional methods and state-of-the-art machine learning baselines across diverse datasets and market conditions.}

\item{We offer new insights into the impact of ensembling on portfolio stability and returns, showing that larger ensembles lead to more robust and reliable performance.}

\end{itemize}

Overall, DSL offers a stable, scalable, and practically effective approach to systematic portfolio optimization, narrowing the gap between academic research and real-world application.

\section{Related Works}

Recent research has demonstrated the promise of machine learning techniques in portfolio optimization, particularly in scenarios with a limited number of assets \cite{lee2024overview}. Gradient-based methods for Sharpe ratio maximization have been explored \cite{Zhang_2020}, while reinforcement learning approaches remain popular and have shown competitive performance in portfolio management tasks \cite{BETANCOURT2021114002, CHAOUKI202016, Jang2023DeepRL, Jiang2024DeepRL, KORATAMADDI2021848, Yu2019ModelbasedDR, acero2024deepreinforcementlearningmeanvariance, SoodDeepRL}. More recently, \cite{lin2024a} proposed a convex optimization framework for Sharpe ratio maximization, further advancing the mathematical tractability of the problem. Large Language Model-based strategies have also proposed \cite{kim2025guruagents, lee2025llmenhanced}.

While classical mean--variance optimization provides a theoretically appealing foundation for portfolio construction, it is sensitive to estimation errors. Recent works, including simulation-based extensions, aim to improve the stability of asset allocation \cite{kim2021mean, chung2022effects}. 

\textit{Decision-Focused Learning} (DFL) has attracted growing interest as an alternative to traditional \textit{Prediction-Focused Learning} (also known as `predict-then-optimize') in portfolio optimization. DFL directly integrates decision-making objectives into the training process and has been theoretically analyzed and empirically compared to predictive approaches \cite{Mandi_2024}. Its successful applications span various domains, including combinatorial optimization and resource allocation \cite{10.1007/978-3-031-44505-7_34, NEURIPS2022_0904c7ed, kong2023df2distributionfreedecisionfocusedlearning}. In the context of financial modeling, \cite{lee2024anatomymachinesmarkowitzdecisionfocused} introduced a DFL-based objective tailored for portfolio optimization. 

\textit{End-to-end learning} for portfolio optimization, a closely related paradigm with DFL, has gained interest in recent years. Approaches such as \cite{zhang2021universalendtoendapproachportfolio} and \cite{uysal2021endtoendriskbudgetingportfolio} demonstrate the effectiveness of directly optimizing allocation rules via neural networks, circumventing the limitations of mean-variance optimization. Advances in DFL and E2E highlight a shift from traditional two-stage modeling to approaches that explicitly consider downstream optimization during training.

Recent advances in deep sequence modeling, such as LSTM \cite{10.1162/neco.1997.9.8.1735} and Transformer (TRF) \cite{NIPS2017_3f5ee243} architectures, have empowered portfolio optimization models. More recently, the Mamba model has emerged as a linear-time alternative to transformers, enabling scalable sequence modeling with reduced computational overhead \cite{gu2024mambalineartimesequencemodeling}.

Another important line of research explores ensemble methods for both predictive accuracy and robustness. \cite{gupta2022ensembles} provided theoretical insights into the effectiveness of ensembling for classification tasks. Deep ensemble approaches, as studied by \cite{NIPS2017_9ef2ed4b}, have been shown to significantly improve uncertainty estimation and model stability. These advantages are particularly valuable in portfolio optimization, where variance reduction and robust allocation are critical.

Despite these advances, existing approaches often struggle to balance practical robustness, computational tractability, and direct alignment with financial performance metrics. Our work addresses these challenges by proposing a Decision by Supervised Learning (DSL) framework that leverages convex surrogate losses and deep ensembling, delivering both stable and superior out-of-sample performance across diverse asset universes.

\section{Method}
\subsection{Preliminaries: Prediction-Focused Learning, Decision-Focused Learning, and End-to-End Learning}

Prediction-Focused Learning (PFL), also known as predict-then-optimize or two-stage learning, addresses decision-making problems by first predicting relevant parameters (such as expected returns and covariances) and then optimizing a decision (e.g., portfolio weights) in a separate step. Formally, portfolio optimization under PFL can be represented as follows:

\begin{equation}
\text{Stage 1 objective: } \hat{\theta} = \argmin_\theta \mathcal{L}(\hat{y}, y), 
\quad \hat{y} = f_\theta(x)
\end{equation}

\begin{equation}
\text{Stage 2 objective: } \hat{w} = \argmax_{w} \mathcal{P}(w, \hat{y})
\end{equation}

where $\mathcal{L}$ is a standard loss function (e.g., mean squared error), $\theta$ denotes model parameters, $y$ the ground-truth statistics, $\mathcal{P}$ a portfolio performance metric (e.g., Sharpe ratio), and $w$ the portfolio weights. Additional constraints can be imposed at the optimization.

However, this approach optimizes prediction accuracy, which does not always guarantee optimal decisions for the downstream task. Decision-Focused Learning (DFL) addresses this disconnect by directly optimizing the ultimate decision quality, thereby aligning the learning process with the final objective. In DFL, the two-stage process is unified by incorporating the decision-making procedure into the learning objective:

\begin{equation}
\text{Objective: } \hat{\theta} = \argmax_\theta \mathcal{P}(w^*(\hat{y}), y), \quad \hat{y} = f_\theta(x)
\end{equation}

where $w^*$ denotes an optimization operator (e.g., a differentiable optimization layer) that outputs the optimal decision based on model predictions. This direct optimization ensures that the learned representations are tailored to improve decision-making performance, not just predictive accuracy.

End-to-End Learning shares a similar philosophy in that model training is directly driven by the final task objective. However, the fundamental distinction is that End-to-End Learning typically uses the model output as the decision itself, bypassing a separate optimization layer:

\begin{equation}
\label{E2E}
\text{Objective: } \hat{\theta} = \argmax_\theta \mathcal{P}(\hat{w}, y), 
\quad \hat{w} = f_\theta(x)
\end{equation}

While this is a simpler alternative to DFL, it may not be sufficiently expressive for problems where an explicit optimization step is crucial.

\subsection{Preliminaries: Deep Ensemble}

Deep Ensembles enhance predictive reliability and accuracy by aggregating outputs from multiple independently trained neural networks. Each network is initialized with a different random seed, promoting diversity in their learned representations. This diversity is the primary driver behind Deep Ensemble's empirical success, leading to improved accuracy, better-calibrated uncertainty estimates, and increased robustness to overfitting and noise.

From a theoretical perspective, the main benefit of Deep Ensembles is variance reduction, analogous to the effect of Bootstrap Aggregation (Bagging) \cite[Section~6.3.1]{10.5555/3217448}. Consider an ensemble of $m$ models, $f_1, f_2, \ldots, f_m$, with the ensemble prediction defined as:

\begin{equation}
\bar{f}(x) = \frac{1}{m} \sum_{i=1}^{m} f_i(x)
\end{equation}

Suppose the variance of each model’s prediction is $\mathbf{Var}[f_i(x)] = \sigma^2$, and the covariance between predictions of different models is $\mathbf{Cov}[f_i, f_j] = \rho \sigma^2$ for $i \neq j$. Then, the variance of the ensemble prediction becomes:

\begin{equation}
\mathbf{Var} \bigl[ \bar{f}(x)\bigr] = \frac{\sigma^2 \bigl(1 + (m-1)\rho\bigr)}{m}
\end{equation}

As long as the pairwise correlation $\rho$ is less than one, which holds when Deep Learning models are initialized and trained independently, the ensemble's variance is strictly less than that of a single model. This property applies to both regression and classification settings, making Deep Ensembles a generally powerful tool for improving the robustness and reliability of deep learning systems.

\subsection{Decision by Supervised Learning with Deep Ensemble}

We introduce Decision by Supervised Learning (DSL), a framework inspired by End-to-End Learning. DSL formulates decision-making as a supervised learning task: instead of predicting intermediate quantities, the model directly predicts optimal decisions (e.g., portfolio weights), and is trained using a cross-entropy loss with respect to precomputed, pre-computed target portfolios.

Unlike previous works such as \cite{Zhang_2020}, we employ a cross-entropy loss, which stabilizes training and enables robust optimization. Nevertheless, the inherent randomness from parameter initialization can introduce variability into individual deep learning models, thereby affecting decision quality. To mitigate this, we leverage Deep Ensembles: by averaging the predictions of multiple independently trained models, we obtain ensemble portfolio weights that are less sensitive to the idiosyncrasies of any single model.

Mathematically, while our approach is philosophically aligned with End-to-End Learning, its formulation slightly differs from \eqref{E2E}:

\begin{equation}
    \{\hat{\theta}_i\}_{i=1}^m = \argmax_{\{\theta_i\}_{i=1}^m} \; \mathcal{L}(\hat{w}, w^*), 
    \quad \text{where} \quad 
    \hat{w} = \frac{1}{m} \sum_{i=1}^{m} f_{\theta_i}(x)
\end{equation}

where $f_1, f_2, \ldots, f_m$ denote the independently initialized models within the Deep Ensemble, and $w^*$ is the hypothetically optimal portfolio used as the supervised target. We explain the details about the objective we used in our experiments at section \ref{setup}.

In summary, our DSL with Deep Ensemble approach combines the strengths of robust aggregation and decision-driven training objectives, providing a practical yet principled framework for portfolio optimization under uncertainty.

\section{Experiment}

In this section, we describe how we evaluated the performance and robustness of our proposed framework, DSL, using backtesting on real-world financial time series. Our experiments are designed to answer the following key research questions:
\begin{itemize}
\item{RQ1. Does DSL deliver superior portfolio performance compared to traditional methods and machine learning baselines?}
\item{RQ2. How does DSL perform across different market universes, model architectures, and portfolio objectives?}
\item{RQ3. What is the impact of ensemble size on the stability and effectiveness of portfolio decisions?}
\end{itemize}

To address these questions, we conduct extensive backtesting using a variety of static and rolling stock universes, comparing cumulative return, Sharpe ratio, and Sortino ratio. We also analyze the effect of Deep Ensembles to understand their contribution to model stability.

\subsection{Experimental setup}
\label{setup}

\textbf{Backtesting and Baselines.} We conduct extensive backtesting of our proposed framework across multiple investment universes, benchmarking its performance against established baselines. Specifically, we compare DSL to two widely-used reference methods: Prediction-Focused Learning (PFL) and End-to-End Learning (E2E), as described in \cite{Zhang_2020}. Every variations we experiment is summarized in Table \ref{tab:settings}. We also report the performance of classical baselines, equal-weighted, value-weighted and Portfolio Optimization method from \cite{lin2024a}. \cite{lin2024a} propose a portfolio optimization method utilizing proximal gradient ascent which maximize the Sharpe Ratio while limiting the number of items to invest to $m$. We set $m$ as number of total items in the universe, thus not imposing any constraints to maximize backtesting performances.

\textbf{Prediction-Focused Learning (PFL).} PFL represents the traditional approach to portfolio optimization, where Deep Sequence models predict future asset returns, followed by a separate optimization step—typically using mean-variance methods. In our implementation, PFL predicts one-month-ahead returns, performs portfolio optimization accordingly.

\textbf{End-to-End Learning (E2E) Extension.} E2E directly optimizes financial metrics such as the Sharpe ratio \cite{RePEc:ucp:jnlbus:v:39:y:1965:p:119} and Sortino ratio \cite{Sortino1994PerformanceMI}, and closely aligns with the Decision-Focused Learning paradigm. We apply the method originally proposed in \cite{Zhang_2020} as E2E baseline. In E2E, the model predicts portfolio weights directly, and performance is evaluated using realized returns. We only experiment E2E instead of DFL since their theoretical background is similar and E2E is a simpler alternative to DFL.

\textbf{Evaluation Design and Metrics.} We evaluate the comparative effectiveness of PFL, E2E, and our Decision by Supervised Learning (DSL) framework using three architectures—Mamba, LSTM, and TRF—across two portfolio optimization objectives: Maximum Sharpe Ratio and Maximum Sortino Ratio. Evaluation metrics include cumulative returns, Sharpe ratios, and Sortino ratios. We omit explicit reporting of Maximum Drawdown (MDD), as it can be roughly inferred from the backtesting visualizations.

\begin{table}[ht]
  \caption{Overview of Experiment Configurations}
  \label{tab:settings}
  \centering
  \begin{tabular}{lll}
    \toprule
    \textbf{Category} & \textbf{Variation} & \textbf{Abbreviation} \\
    \midrule
    \multirow{3}{*}{Performance} 
      & Cumulative Return           & CR \\
      & Sharpe Ratio                & SH \\
      & Sortino Ratio               & SO \\
    \midrule
    \multirow{3}{*}{Benchmark} 
      & Equal-weighted              & EW \\
      & Value-weighted              & VW \\
      & Portfolio Optimization \cite{lin2024a} & mSSRM \\
    \midrule
    \multirow{2}{*}{Objective}
      & Maximum Sharpe Ratio        & MSH \\
      & Maximum Sortino Ratio       & MSO \\
    \midrule
    \multirow{3}{*}{Model}
      & Mamba                       & M \\
      & LSTM                        & L \\
      & Transformer                 & T \\
    \midrule
    \multirow{3}{*}{Method}
      & Prediction-Focused Learning & PFL \\
      & End-to-End Learning \cite{Zhang_2020} & E2E \\
      & Decision by Supervised Learning & DSL (Ours) \\
    \bottomrule
  \end{tabular}
\end{table}

\textbf{Input Data and Preprocessing.} The input data for PFL, E2E, and DSL consists of preprocessed OHLCV (Open, High, Low, Close, Volume) financial time series, transformed using log-return formulas. Specifically, `Open', `High', and `Low' are each divided by the corresponding `Close' value and then log-transformed; `Close' is divided by its previous day's value and log-transformed; and `Volume' is log-transformed directly. Each input sample to the Deep Sequence models comprises a sequence of 21 consecutive trading days of transformed data. The training period begins in 2010, while the backtesting period starts in November 2019 to include the period of the significant market volatility caused by COVID-19.

\textbf{Target Portfolio Construction.} The target for our supervised learning model is a pre-computed, hypothetically optimal portfolio, which is updated monthly. To construct Maximum Sharpe Ratio and Maximum Sortino Ratio portfolios, we use the convexification approach of \cite{Dinkelbach1967OnNF}. The lookback period for calculating the target portfolio return each month is typically set to 21 trading days.

\textbf{Training and Ensemble Evaluation.} We use Sophia optimizer \cite{liu2024sophiascalablestochasticsecondorder} in every experiments due to its computational efficiency. For every learning-based strategies (PFL, E2E and DSL), we adopt a rolling prediction approach, retraining the models each month for 100 epochs and selecting the epoch with the lowest validation loss. The final month of each training window is reserved for validation. To enhance robustness and stability, DSL constructs portfolio allocations by averaging the weights from 100 independent runs (Deep Ensemble). In contrast, for both PFL and E2E, we report performance metrics as averages over 100 independent runs.

\subsection{Evaluation}

We evaluate our framework and all baseline methods across eight distinct investment universes. Four universes (Large Cap, Range-bounded, High Dividend, Utility Sector) are constructed through manual selection of constituent stocks, while the remaining universes (NASDAQ Top 30, S\&P Top 30) are defined by objective criteria, such as the current top 30 constituents of major market indices. To further challenge learning-based methods, we also conduct experiments in rolling universes (NASDAQ Rolling S\&P Rolling), where index membership is updated monthly—an especially rigorous and dynamic setting. We collect and utilize the data from January 2010 to May 2025.  Details of all universe compositions are provided in our linked anonymized GitHub repository. While the raw datasets cannot be shared due to licensing constraints, we make all experimental results and code publicly available.

To ensure robust and reliable backtesting results, we have implemented rigorous controls to address common challenges such as look-ahead bias, survivorship bias, and backtest overfitting. Specifically, our framework incorporates a two-day trading lag and applies realistic transaction costs of 0.4\% per trade, both for buying and selling. All predictive models operate strictly on data that would have been available at the time of decision-making, thereby eliminating any possibility of look-ahead bias. Furthermore, we set the risk-free rate to zero within our backtesting environment for simplicity and transparency. While this assumption may introduce minor differences compared to certain real-world scenarios, it allows us to focus on the core performance and risk characteristics of our strategies.

\begin{table*}[h]
\caption{Backtesting results across static universes.}
\label{tab:backtest-onerow1}
\small
\begin{tabular}{lccccccccccccccccccccccccc}
\toprule
& \multicolumn{3}{c}{\textbf{Large Cap}} & \multicolumn{3}{c}{\textbf{Range-Bounded}} & \multicolumn{3}{c}{\textbf{High Dividend}} & \multicolumn{3}{c}{\textbf{Utility Sector}} & \multicolumn{3}{c}{\textbf{NASDAQ Top 30}} & \multicolumn{3}{c}{\textbf{S\&P Top 30}} \\
\textbf{Setting} & \textbf{CR} & \textbf{SH} & \textbf{SO} & \textbf{CR} & \textbf{SH} & \textbf{SO} & \textbf{CR} & \textbf{SH} & \textbf{SO} & \textbf{CR} & \textbf{SH} & \textbf{SO}
& \textbf{CR} & \textbf{SH} & \textbf{SO} & \textbf{CR} & \textbf{SH} & \textbf{SO}  \\
\midrule
EW & 2.79 & 0.78 & 1.21 & 1.79 & 0.41 & 0.64 & 1.30 & 0.28 & 0.42 & 1.27 & 0.20 & 0.30 & 3.41 & 0.86 & 1.36 & 2.64 & 0.90 & 1.37 \\
VW & 4.61 & 0.79 & 1.25 & 2.02 & 0.43 & 0.68 & 1.24 & 0.23 & 0.35 & 1.35 & 0.24 & 0.37 & 6.08 & 0.94 & 1.50 & 4.83 & 0.91 & 1.44 \\
mSSRM & 2.97 & 0.76 & 1.24 & 2.22 & 0.56 & 0.86 & 0.90 & -0.09 & -0.14 & 1.08 & 0.05 & 0.08 & 3.17 & 0.88 & 1.40 & 2.57 & 0.67 & 0.96  \\
\midrule
MSH\_M\_PFL & 1.85 & 0.49 & 0.76 & 1.50 & 0.29 & 0.44 & 0.82 & -0.20 & -0.31 & 0.94 & -0.04 & -0.07 & \f{2.37} & \f{0.63} & \f{0.99} & 1.38 & 0.32 & 0.48 \\
MSH\_M\_E2E & 1.50 & 0.33 & 0.51 & 1.65 & 0.31 & 0.50 & 0.80 & -0.21 & -0.32 & \f{0.97} & \f{-0.02} & \f{-0.03} & 2.29 & 0.56 & 0.89 & 1.81 & 0.56 & 0.85 \\
MSH\_M\_DSL (Ours) & \f{2.68} & \f{0.78} & \f{1.19} & \f{1.66} & \f{0.36} & \f{0.56} & \f{1.50} & \f{0.39} & \f{0.59} & 0.96 & -0.02 & -0.04 & 2.05 & 0.52 & 0.80 & \f{2.80} & \f{0.91} & \f{1.43} \\
\midrule
MSH\_L\_PFL & 2.37 & 0.65 & 1.01 & 1.48 & 0.28 & 0.43 & 0.85 & -0.16 & -0.24 & 0.91 & -0.06 & -0.10  & \f{2.29} & \f{0.59} & \f{0.93} & 1.71 & 0.50 & 0.75 \\
MSH\_L\_E2E & 1.61 & 0.38 & 0.60 & \f{1.78} & \f{0.37} & \f{0.59} & 0.58 & -0.46 & -0.71 & \f{0.92} & \f{-0.04} & \f{-0.07} & 1.48 & 0.22 & 0.34 & 1.65 & 0.39 & 0.61 \\
MSH\_L\_DSL (Ours) & \f{2.40} & \f{0.72} & \f{1.11} & 1.38 & 0.21 & 0.34 & \f{2.13} & \f{0.72} & \f{1.11} & 0.74 & -0.21 & -0.31  & 1.61 & 0.39 & 0.61 & \f{2.69} & \f{0.88} & \f{1.35} \\
\midrule    
MSH\_T\_PFL & 1.80 & 0.46 & 0.71 & 1.42 & 0.25 & 0.39 & 0.83 & -0.18 & -0.28 & \f{0.89} & \f{-0.09} & \f{-0.13} & \f{2.46} & \f{0.66} & \f{1.04} & 1.67 & 0.51 & 0.77 \\
MSH\_T\_E2E & 1.56 & 0.36 & 0.57 & \f{1.79} & \f{0.38} & \f{0.59} & 0.64 & -0.39 & -0.59 & 0.82 & -0.10 & -0.14 & 1.62 & 0.27 & 0.42 & 1.70 & 0.41 & 0.65 \\
MSH\_T\_DSL (Ours) & \f{3.05} & \f{0.86} & \f{1.35} & 1.46 & 0.25 & 0.40 & \f{2.08} & \f{0.73} & \f{1.13} & 0.77 & -0.18 & -0.26  & 1.68 & 0.41 & 0.63 & \f{2.92} & \f{0.91} & \f{1.40}  \\
\midrule   
MSO\_M\_PFL & 1.61 & 0.34 & 0.52 & 1.35 & 0.21 & 0.33 & 0.79 & -0.23 & -0.34 & 0.77 & -0.20 & -0.29  & 3.22 & 0.80 & 1.25 & 2.12 & 0.66 & 1.01 \\
MSO\_M\_E2E & 1.45 & 0.29 & 0.45 & 1.38 & 0.20 & 0.31 & 0.80 & -0.23 & -0.35 & 1.04 & 0.03 & 0.04 & 2.07 & 0.52 & 0.83 & 1.68 & 0.49 & 0.74 \\
MSO\_M\_DSL (Ours) & \f{3.60} & \f{0.68} & \f{1.06} & \f{1.56} & \f{0.27} & \f{0.43} & \f{1.75} & \f{0.52} & \f{0.82} & \f{1.05} & \f{0.03} & \f{0.05} & \f{8.38} & \f{0.97} & \f{1.49} & \f{10.14} & \f{1.15} & \f{1.82}\\
\midrule    
MSO\_L\_PFL & 1.97 & 0.49 & 0.75 & 1.34 & 0.20 & 0.32 & 0.83 & -0.18 & -0.26 & 0.82 & -0.15 & -0.22 & 3.22 & 0.78 & 1.23 & 2.00 & 0.62 & 0.94 \\
MSO\_L\_E2E & 1.41 & 0.28 & 0.45 & 1.03 & 0.02 & 0.04 & 0.67 & -0.35 & -0.54 & \f{0.84} & \f{-0.10} & \f{-0.15} & 1.60 & 0.29 & 0.45 & 1.55 & 0.39 & 0.60 \\
MSO\_L\_DSL (Ours) & \f{6.38} & \f{1.02} & \f{1.61} & \f{1.36} & \f{0.17} & \f{0.28} & \f{1.23} & \f{0.19} & \f{0.29} & 0.68 & -0.24 & -0.36 & \f{11.62} & \f{1.04} & \f{1.63} & \f{8.50} & \f{1.10} & \f{1.78} \\
\midrule
MSO\_T\_PFL & 1.82 & 0.45 & 0.68 & 1.30 & 0.19 & 0.29 & 0.82 & -0.20 & -0.29 & \f{0.81} & \f{-0.16} & \f{-0.23} & 3.28 & 0.80 & 1.27 & 2.30 & 0.74 & 1.13 \\
MSO\_T\_E2E & 1.52 & 0.36 & 0.56 & 1.05 & 0.03 & 0.06 & 0.77 & -0.24 & -0.37 & 0.73 & -0.17 & -0.25 & 2.08 & 0.44 & 0.68 & 1.63 & 0.44 & 0.69 \\
MSO\_T\_DSL (Ours) & \f{5.30} & \f{0.82} & \f{1.29} & \f{1.49} & \f{0.23} & \f{0.37} & \f{1.22} & \f{0.18} & \f{0.27} & 0.70 & -0.24 & -0.35 & \f{9.25} & \f{1.01} & \f{1.58} & \f{7.01} & \f{0.94} & \f{1.49} \\
\bottomrule
\end{tabular}
\end{table*}

\begin{table}[h]
\caption{Backtesting results on rolling universes.}
\label{tab:backtest-onerow2}
\small
\begin{tabular}{lcccccc}
\toprule
& \multicolumn{3}{c}{\textbf{NASDAQ 100}} & \multicolumn{3}{c}{\textbf{S\&P 500}} \\
& \textbf{CR} & \textbf{SH} & \textbf{SO} & \textbf{CR} & \textbf{SH} & \textbf{SO} \\
\midrule
EW & 2.05 & 0.54 & 0.85 & 1.60 & 0.39 & 0.58 \\
VW & 2.75 & 0.58 & 0.91 & 1.88 & 0.45 & 0.68 \\
mSSRM & 2.54 & 0.82 & 1.30 & 1.58 & 0.52 & 0.80 \\
\midrule
MSH\_M\_PFL & \f{2.05} & \f{0.55} & \f{0.85} & \f{1.60} & \f{0.39} & \f{0.58} \\
MSH\_M\_E2E & 0.97 & -0.02 & -0.03 & 1.24 & 0.17 & 0.26 \\
MSH\_M\_DSL (Ours) & 1.58 & 0.37 & 0.58 & 1.47 & 0.33 & 0.49 \\
\midrule
MSH\_L\_PFL & \f{1.58} & \f{0.33} & \f{0.51} & 1.32 & 0.22 & 0.33 \\
MSH\_L\_E2E & 1.45 & 0.25 & 0.39 & 0.74 & -0.23 & -0.34 \\
MSH\_L\_DSL (Ours) & 0.97 & -0.01 & -0.02 & \f{1.79} & \f{0.41} & \f{0.63} \\
\midrule 
MSH\_T\_PFL & 1.37 & 0.21 & 0.32 & 1.41 & 0.28 & 0.42 \\
MSH\_T\_E2E & \f{1.48} & \f{0.26} & \f{0.41} & 0.72 & -0.27 & -0.39 \\
MSH\_T\_DSL (Ours) & 1.02 & 0.01 & 0.02 & \f{1.90} & \f{0.45} & \f{0.67} \\
\midrule
MSO\_M\_PFL & 2.05 & \f{0.54} & \f{0.85} & 1.60 & 0.39 & 0.58 \\
MSO\_M\_E2E & 0.99 & -0.00 & -0.00 & 1.37 & 0.23 & 0.36 \\
MSO\_M\_DSL (Ours) & \f{2.14} & 0.49 & 0.75 & \f{1.72} & \f{0.47} & \f{0.69} \\
\midrule
MSO\_L\_PFL & \f{1.85} & \f{0.44} & \f{0.68} & 1.25 & 0.18 & 0.26 \\
MSO\_L\_E2E & 0.70 & -0.23 & -0.35 & 0.72 & -0.29 & -0.45 \\
MSO\_L\_DSL (Ours) & 1.82 & 0.27 & 0.42 & \f{1.64} & \f{0.29} & \f{0.45} \\
\midrule
MSO\_T\_PFL & 1.38 & 0.21 & 0.33 & 1.34 & 0.24 & 0.35 \\
MSO\_T\_E2E & 0.82 & -0.12 & -0.18 & 0.70 & -0.32 & -0.49 \\
MSO\_T\_DSL (Ours) & \f{2.00}& \f{0.28} & \f{0.42} & \f{2.24} & \f{0.47} & \f{0.75} \\
\bottomrule
\end{tabular}
\end{table}

\subsection{Backtesting Experiment (RQ1, RQ2)}

\begin{figure*}[h]
  \centering
  \includegraphics[width=\textwidth]{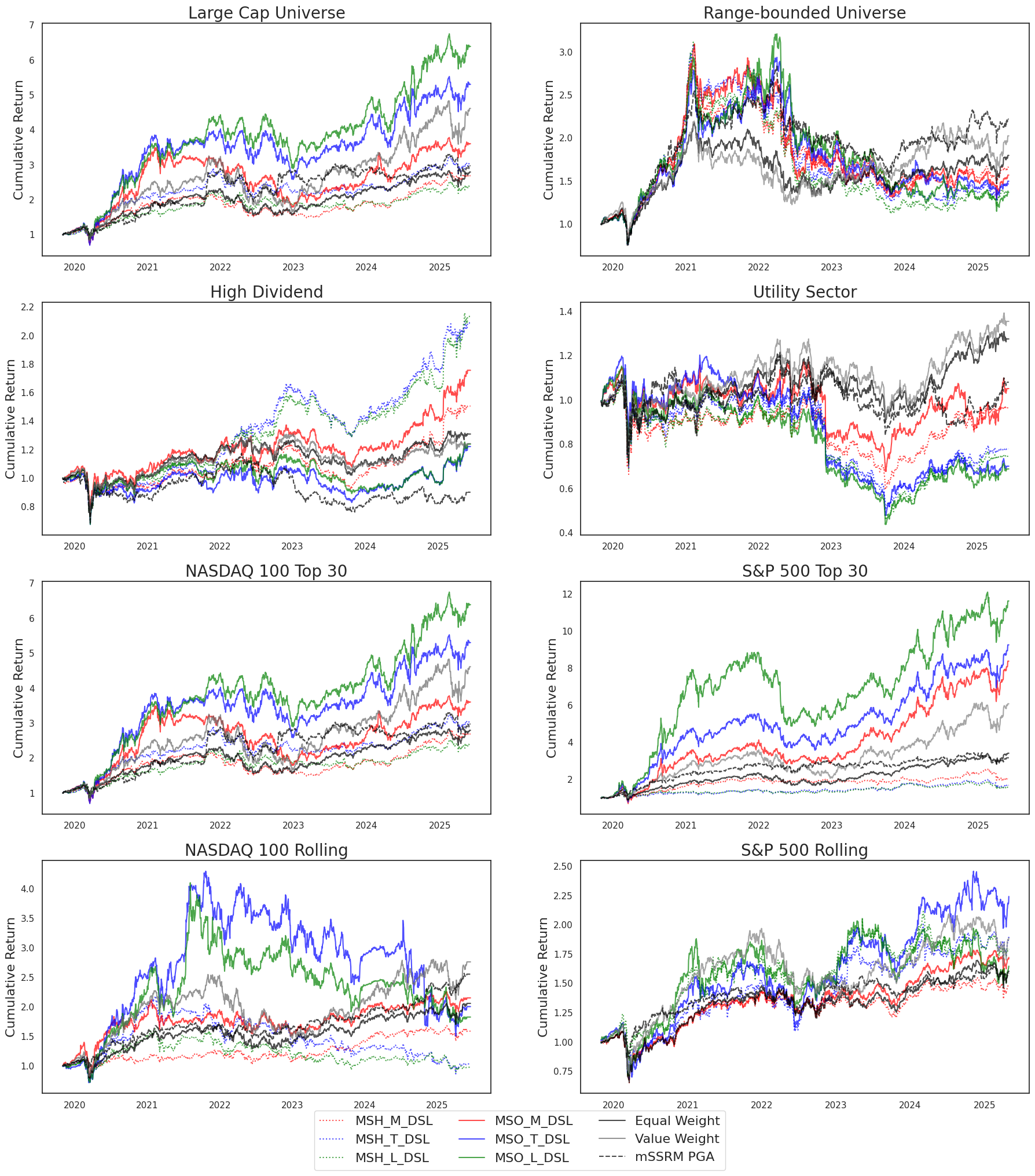}
  \caption{Comparing cumulative return trajectory with classical portfolio strategies.}
  \label{fig:backtest}
  \Description{We compare the backtesting results with classical portfolio strategies, such as equal-weighted, value-weighted portfolios and mSSRM. We visualize cumulative return of different strategies for each universe.}
\end{figure*}

Table \ref{tab:backtest-onerow1} and Table \ref{tab:backtest-onerow2} summarize DSL's backtest results. We report cumulative return, Sharpe ratio, and Sortino ratio for each method. The best performance among DFL, E2E, and DSL is highlighted in bold. In numerous scenario and market universe, DSL outperforms both Machine Learning baselines (PFL and E2E) as well as classical strategies (equal-weight, value-weight and mSSRM), achieving higher cumulative returns and better Sharpe and Sortino ratios.

The Large-Cap and High-Dividend universes (24 and 84 U.S. stocks, respectively) are fixed sets. DSL achieves the best performance across all tested objectives, demonstrating its effectiveness when the stock universe is stable. The Range-bounded Universe and Utility Sector Universe serve as stress tests. The Range-Bounded universe imposes tight price ceilings and floors, yet DSL surpasses PFL and E2E in four out of six cases. In the Utilities universe, which experienced a downtrend, DSL leads in only one of six cases—suggesting the method can behave similarly to a leveraged strategy under prolonged declines. Table \ref{tab:backtest-onerow1} also covers more familiar index-based universes. In static NASDAQ Top 30 and S\&P Top 30 universes, DSL again leads, except for the Max-Sharpe objective in NASDAQ. 

To address potential survivorship bias in these fixed universes, we also evaluate rolling versions of the NASDAQ 100 and S\&P 500, updating stock lists each month to reflect actual index membership. Table \ref{tab:backtest-onerow2} represent two rolling universes. Even in these more challenging and dynamic settings, DSL maintains a competitive edge, especially in the S\&P 500 universe. DSL reports best Cumulative Return, Sharpe Ratio and Sortino Ratio in 5 experimental settings among 6 configuration within this universe.  

Figure \ref{fig:backtest} compares DSL (with different objectives and models) to traditional portfolio strategies across each universe. The results indicate that, in many cases, DSL's choice of objective and model outperforms conventional benchmarks. We can observe that in most cases, Maximum Sortino Ratio strategies shows outperformance over baselines, while Maximum Sharpe Ratio strategies shows underperformance over baselines. Thus, selecting appropriate objective is important for the portfolio optimization performance within our framework.

Overall, DSL delivers strong and consistent results across diverse market conditions, establishing itself as a practical and systematic tool for investors aiming to build superior portfolios.

\subsection{Ensemble Effect Analysis (RQ3)}

\begin{figure}[h]
  \centering
  \includegraphics[width=\columnwidth]{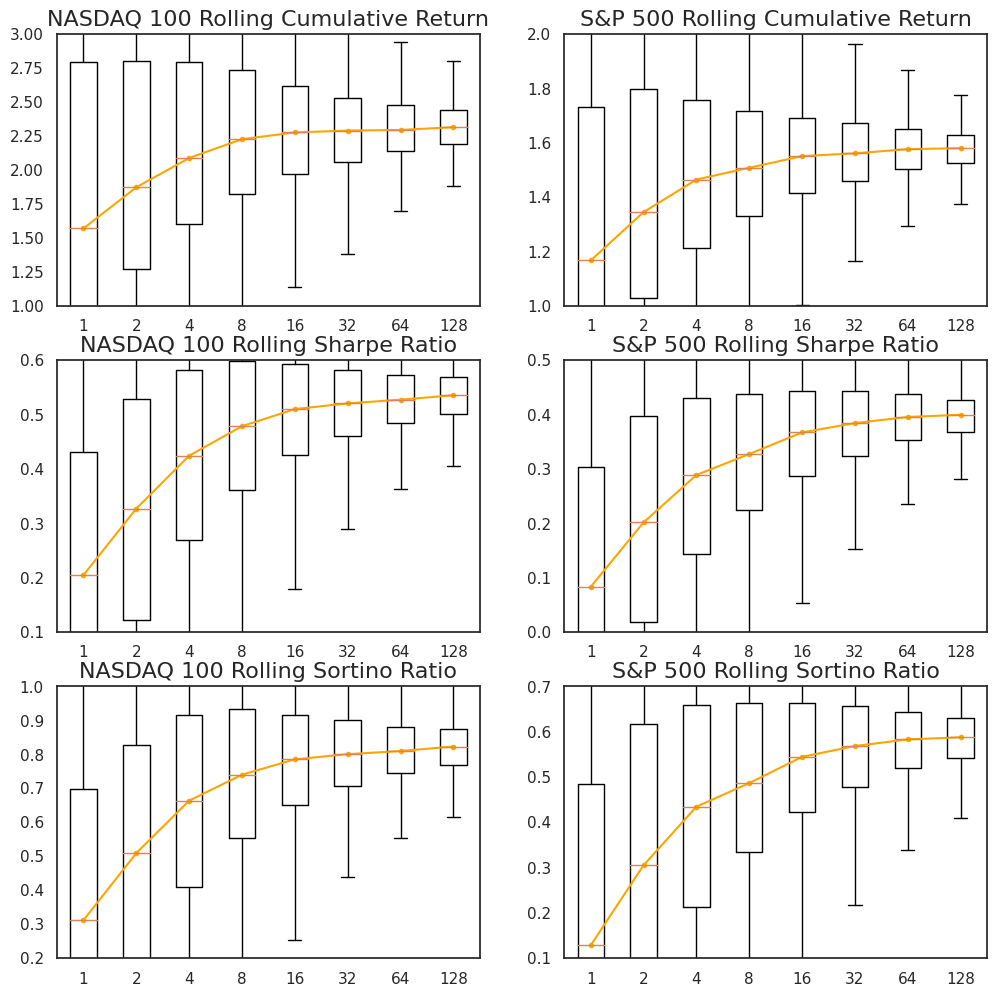}
  \caption{Impact of Ensemble Size on Performance}
  \label{fig:ensemble}
  \Description{Ensemble effect experiment on NASDAQ 100 Rolling universe and S\&P500 Rolling universe}
\end{figure}

We investigate the impact of the Deep Ensemble method within our experimental framework, focusing on its effectiveness in portfolio optimization. For this analysis, we select the NASDAQ 100 Rolling Universe and S\&P 500 Rolling Universe as our investment universes. The experiments center on the Mamba model under the Maximum Sortino Ratio target portfolio scheme, chosen for its consistently strong backtesting performance, low computational cost, and relatively high variability—making it an ideal subject for studying ensemble effects.

Figure \ref{fig:ensemble} illustrates how the Deep Ensemble approach enhances portfolio optimization for the Mamba model. To rigorously assess ensembling, we generate 1,000 distinct model samples and use bootstrapping to compute ensemble weights. For each ensemble size, we repeatedly sample models with replacement (1,000 iterations), average their portfolio weights, and evaluate the backtested performance of the aggregated weights. The median in each box plot is indicated by the orange line. As ensemble size increases from 1 to 64, cumulative returns steadily improve, clearly demonstrating the power of ensembling to reduce variability and enhance returns. Both the Sharpe and Sortino ratios exhibit a strong upward trend, while the narrowing interquartile ranges in the box plots reveal that larger ensemble sizes substantially reduce variance and improve the stability of portfolio estimates.

\section{Limitations}

While the DSL framework with Deep Ensembles demonstrates strong empirical performance and practical robustness, several limitations remain that merit consideration and open avenues for further improvement.

\textbf{Model and Loss Function Constraints}. Our framework currently employs cross-entropy loss and is designed for long-only portfolios. This restricts its direct applicability to scenarios where short-selling is required or portfolio constraints are more complex. Although the framework could be extended by adopting alternative loss functions (such as Mean Squared Error for long-short portfolios) or custom constraints, this would require careful design and additional validation.

\textbf{Computational Overhead}. Deep Ensemble methods significantly increase computational costs, as they require training and maintaining multiple independent models. While parallelization and modern computing resources can alleviate some of this burden, practitioners must carefully balance ensemble size and performance gains against available resources, especially in time-sensitive or high-frequency trading environments.

\textbf{Sensitivity to Market Regimes}. Although DSL generally outperforms baselines across a variety of datasets, our experiments reveal that its advantage may diminish or even reverse in certain adverse market conditions (e.g., range-bounded, or downward trending). In such cases, DSL can behave as a high-beta strategy, potentially exposing investors to greater downside risk. Practitioners should therefore consider additional risk management overlays or adaptive objective functions to mitigate this sensitivity.

\section{Conclusion}

We have introduced Decision by Supervised Learning (DSL), a novel and practical portfolio optimization framework that adapts End-to-End Learning to robustly address real-world investment problems. DSL reframes portfolio construction as a supervised learning task, leveraging target portfolios such as Max-Sharpe and Max-Sortino and optimizing a cross-entropy loss function. To further improve performance and stability, we incorporate Deep Ensemble methods, effectively reducing variance and enhancing the reliability of portfolio allocations.

Comprehensive experiments on diverse market universes demonstrate that shows superior performance compared to both traditional portfolio strategies and recent machine learning-based approaches—including Prediction-Focused Learning and End-to-End Learning—across a variety of market conditions. Our analysis of ensemble size reveals that larger ensembles deliver higher median returns and greater stability, making Deep Ensemble a particularly effective tool in practical portfolio management.

Despite these promising results, DSL has some limitations. The current framework is tailored for long-only portfolios, and certain market scenarios may reduce its comparative advantage. Additionally, employing Deep Ensembles increases computational demands, requiring practitioners to balance performance benefits against resource constraints.

Looking forward, several extensions and research directions are possible. Future work could adapt DSL to support long-short portfolios, incorporate additional financial objectives and constraints, or explore alternative loss functions for even greater flexibility. There is also potential to integrate DSL with reinforcement learning or advanced uncertainty estimation techniques for more adaptive and resilient strategies.

Overall, DSL with Deep Ensembles offers a practical, robust, and extensible solution for systematic portfolio optimization. We believe this framework has strong potential to benefit both researchers and practitioners in the evolving field of quantitative finance.

\begin{acks}
This research was supported by \textbf{Brian Impact Foundation}, a non-profit organization dedicated to the advancement of science and technology for all.

This document was prepared by the \textbf{AI Solution Team} at \textbf{Mirae Asset Global Investments} for informational purposes only. \textbf{Mirae Asset Global Investments} makes no representation or warranty as to the accuracy, completeness, or reliability of the information contained herein and accepts no liability for any consequences arising from its use. This document does not constitute investment advice, a recommendation, or an offer to buy or sell any financial products, and should not be relied upon as a basis for investment decisions.
\end{acks}

\bibliographystyle{ACM-Reference-Format}
\bibliography{references}

\end{document}